\documentclass[11pt]{article}
\usepackage{authblk}
\usepackage{coling2016}
\usepackage{times}
\usepackage{url}
\usepackage{latexsym}
\usepackage{CJK}
\usepackage{graphicx}
\usepackage[T1]{fontenc}
\usepackage[utf8]{inputenc}

\usepackage{multirow}
\newcommand{\tabincell}[2]{\begin{tabular}{@{}#1@{}}#2\end{tabular}}

\title{Learning to Start for Sequence to Sequence Architecture}

\author[*]{Qingfu Zhu}
\author[*]{Weinan Zhang}
\author[**]{Lianqiang Zhou}
\author[*]{Ting Liu}
\affil[*]{Harbin Institute of Technology, \authorcr Email:\{qfzhu,wnzhang,tliu\}@ir.hit.edu.cn}
\affil[**]{Tencent Corporation, \authorcr Email:tomcatzhou@tencent.com}

\date{}

\begin{document}
\maketitle
\begin{CJK*}{UTF8}{gbsn}
\begin{abstract}
The sequence to sequence architecture is widely used in the response generation and neural machine translation to model the potential relationship between two sentences.
It typically consists of two parts: an encoder that reads from the source sentence and a decoder that generates the target sentence word by word according to the encoder's output and the last generated word.
However, it faces to the ``cold start'' problem when generating the first word as there is no previous word to refer.
Existing work mainly use a special start symbol ``$<$/s$>$'' to generate the first word.
An obvious drawback of these work is that there is not a learnable relationship between words and the start symbol.
Furthermore, it may lead to the error accumulation for decoding when the first word is incorrectly generated.
In this paper, we proposed a novel approach to learning to generate the first word in the sequence to sequence architecture rather than using the start symbol.
Experimental results on the task of response generation of short text conversation show that the proposed approach outperforms the state-of-the-art approach in both of the automatic and manual evaluations.
\end{abstract}
\section{Introduction}
Recently, the sequence to sequence(Seq2Seq) architecture has gained great
development as a general neural network method to model the potential
relationship between two sequences.
For the basic Seq2Seq model, each sequence is usually modeled by RNN,
and the two RNNs for the source sequence and target sequence are called encoder and decoder
respectively. The encoder reads from the source sentence
and do some summarize. The decoder is actually a language model that produce words
according to previously predicted words conditioned with the encoder's output(usually called the context
vector). This
indicates that when
the decoder try to predict a word, the context vector and the word predicted at previous time are two necessary inputs that requires.

So here comes the initialization question: when producing the first word by the decoder,
there is no previous predicted word to be referenced to.
Typically, previous work use a start symbol ``$<$/s$>$'' to generation the first word \cite{Sutskever2014Sequence}.
While it is not suitable to introduce a start symbol as the first word varies from different sentences.
Concretely, there is not a learnable conditional probability of words given start symbol.
Meanwhile, the process of producing the first word and generating the rest words of a sentence are different so that they  should be handled respectively.
To address this issue, we proposed a novel approach to learning to generate the first word.
In detail, we find two factors that impact the encoding and decoding process:
one is the source sequence which can be expressed using the encoder's states.
The other is the representation of candidate words, of which information are all contained
in the embedding matrix.
We thus introduce these variables to map the representation of the source sentence into a probability distribution over the word table, pick up the maximal dimension as the final result.

The contribution of this paper is as follows:
\begin{itemize}
  \item To the best of our knowledge, we are the first to proposed a novel approach to learning to generate the first word in Seq2Seq architecture.
  \item The proposed approach outperforms the state-of-the-art on the response generation of the short text conversation.
  \item Besides the short text conversation task, the proposed approach is a general framework which can also adapt to other Seq2Seq learning applications.
\end{itemize}

\section{Background}
From the perspective of probability, the Seq2Seq model maximize the probability
of the target sequence conditioned with the source sequence
during the training process, and search for a sequence that
have a maximal conditioned probability given the source sequence
during the predicting process. Due to that highly abstract attribute,
lots of tasks such as Response Generation,Machine Translation and Question Answering can
all be modeled using that architecture.
\subsection{RNN encoder-decoder}
Typically, a sequence to sequence model consists of two parts: encoder
and decoder, both of which are often implemented using a family
of RNN, such as GRU \cite{Cho2014On,Chung2014Empirical} and LSTM \cite{hochreiter1997long,gers2000learning,graves2012neural}, so a seq2seq model is also called RNN
encoder-decoder architecture.

The encoder is a normal RNN, which reads from a sequence of words
and outputs their hidden states.These states are
also called annotations denoted by $H$,
and for each hidden state $h_i$ at time $i$, it is computed by its
previous hidden state $h_{i-1}$ and the word at time $t$:
\begin{equation}\label{encoder hidden state}
h_t = f(h_{t-1}, x_t); \quad H = \{h_1, h2, h3, ..., h_T\}
\end{equation}
Here, $T$ is the length of the source sequence, $f$ is a non-linear function.
After that, the encoder computes a distributed representation using
these hidden states as a summary(context vector) of the input sequence.
The most simplest way is directly fetching the last one:
\begin{equation}\label{encoder context vector}
c = q(\{h_1, h2, h3, ..., h_T\})
\end{equation}

For the decoder, the hidden state's calculation is quite similar,
the only difference is that the sequence input $x_t$ is replaced
by the word predicted at last time:
\begin{equation}\label{decoder hidden state}
s_t = f(s_{t-1}, y_{t-1})
\end{equation}

It should be noted that, the context vector $c$ is
used to initialize the hidden state of decoder to make sure that the
decoder was conditioned with the encoder. Based on that, \cite{Cho2014Learning} add
the vector $c$ as an extra input into the computation of the hidden
state in decoder to make sure that every time step of the decoder can
get full information of the context. In that way, the formula \ref{decoder hidden state}
should be updated to:
\begin{equation}\label{dec state}
s_t = f(s_{t-1}, y_{t-1}, c)
\end{equation}

Then, the word at time t can be predicted by mapping the $s_t$ to a probability
over the word table using the maxout activation\cite{goodfellow2013maxout}.

\subsection{Attention Mechanism in Seq2Seq}
In the basic architecture of the sequence to sequence, source sequence
sent to the encoder is encoded into a dense, fixed-length
vector. Considering that vector may not be able to contain all the
useful information of the source sequence, thus becoming a bottleneck of the
model,\newcite{bahdanau2014neural} add the attention mechanism to improve the Seq2Seq's
performance.Compared with the basic architecture, which use the last hidden
state as the context vector $c$,
attention mechanism gives a weight to all the annotations, then use them
to calculate a weighted sum as a new context vector. It should be noted that, in that
way, the vector $c$ is distinct for every time step in the decoder, because a time-related
variable was involved during the computing, so here we denote the result as vector $c_j$.
\begin{equation}
c_j = \sum_{i=1}^T \alpha_{ij} h_i
\end{equation}
Here, $c_j$ is the context vector when we decode the $j$-th word in the decoder,
and the weight $\alpha_{ij}$ for the $i$-th annotation of encoder is computed by:
\begin{equation}\label{alpha}
\alpha_{ij} = a(s_{j-1}, h_i)
\end{equation}
where $a$ is a forward neural network. Intuitively, the vector $s_{j-1}$ contains
the context information of the response, so Formula \ref{alpha} can be understanded as to
calculate the similarity between that context and these encoder annotations, which can be also regarded
as a weight.
\subsection{Initialization in Seq2Seq Learning}
Initialization is such a small detail that can be ignored easily,sometimes.
However, it is an important part of the model. In the encoder RNN, a
state will be used to compute the state at next time(see Eq.\ref{encoder hidden state}), and by this way, the
initial state will have an indirectly influence on all the states next. The
decoder RNN share the same situation. In addition, the
decoder has an extra variable that should be initialized: predicted word at last
time step, because we don't have that input for the first process of generation.
Typically, we set the initial hidden state of encoder
 to an all zero vector, and people usually use the last hidden state
of the encoder to initialize the decoder's first hidden states:
\begin{equation}
s_0 = \sigma(W_s h_T)
\end{equation}
where $\sigma$ is a non-linear function, $W_s$ is a trainable parameter. That is intuitively plausible because it describes
the relation between the two sequences that the decoder is conditioned with the
encoder.  As to the previous generated word for first generation in the decoder, we manually
set a start symbol to act as that role. 

\section{Learning to start}

In this section, we propose a new model to accomplish Seq2Seq's the
initial prediction.
We think that the method using a start symbol to predict the first word is not very
suitable. First, the decoder RNN is essentially a language model \cite{mikolov2010recurrent}, which use the previous
predicted words to predict a new word, from the perspective of probability,
it learns a conditional probability of word that given last predicted words.
While the start symbol and the first predicted word do not have such association,
because most words can be put at the first position of a sequence, there is
not a learnable conditional probability, so the result of taking a start symbol may
cause the model prefer to predict some high frequency words, which is also observed
during other conversation models using this architecture \cite{sordoni2015neural,serban2015hierarchical,vinyals2015neural}.
We think the reason may lie in training samples started with these words takes a higher proportion,
making the decoder
learn a conditional probability that given the start symbol, these words' probabilities
should be higher than others. Second, we suppose the process of predicting the first
word and predicting a word according to its previous word should not be treated identically,
using a same method to do the two works may not be a good choice. Third, the start symbol
is involved in the calculation of decoder's hidden
state(see Eq. \ref{dec state}),
so introducing a start symbol that irrelevant of a sequence and
no difference between all the source sequences may indirectly influence the prediction of
the rest time steps.
\begin{figure}[!ht]	
	\centering
    \includegraphics[width=0.4\textwidth,natwidth=610,natheight=642]{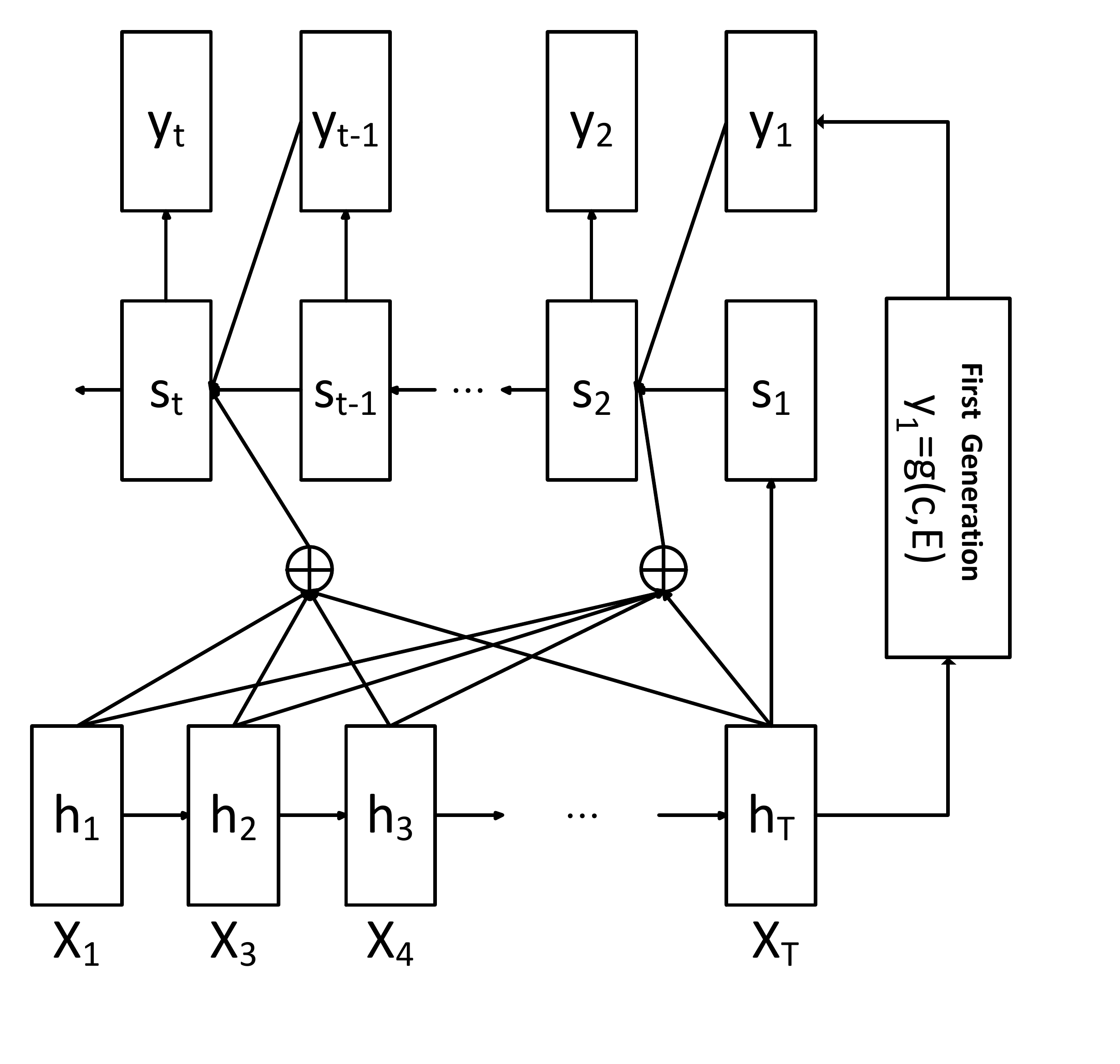}\\
	\caption{The LTS model architecture.}
	\label{LTS}
\end{figure}

So we propose a new method to relive the decoder from both predicting the first word and
predicting word according to the last predicted word. In our model, the first word is
predicted independently from the decoder. Inspired by the initialization of the hidden
state of the decoder, we use the hidden state of encoder to calculate the probability of
the first word using the formula below:
\begin{equation}
y_0 = \sigma((\sigma(W_i c)+b_i)E + b_e)
\end{equation}

In this formula, vector $c$ is the context vector, here we directly use the last
hidden state of the encoder, $W_i$ is a
matrix that can be trained in the model, $E$ is the embedding matrix of the
decoder, and $b_i, b_e$ are bias vectors, $\sigma$ is a no-linear activation function,
so in form, the formula is equal to below if we ignore all the bias:
\begin{equation}
y_0 = g(c, E)
\end{equation}

Intuitively, this formula build a tie between the context vector and the
embedding matrix of the decoder. The former contains information of the source
sequence, and the latter contains information of all the candidate words to be
predicted in the decoder. So the $W_i$ matrix can be regarded as a similarity matrix
used to compute the probability that how similar a word is to the source sequence, which
indicates whether it is suitable to be predicted as the first word. Besides, by doing
like this, the generation of the first word is decided only by the encoder's state. And without a start
symbol's influence, the encoder's state can also be transferred to the decoder without any
loss. And the rest process of prediction remains the same to the basic structure.

\section{Experiment Settings}
To verify the effectiveness, our proposed model were tested in the task of response generation of short text conversation. As a kind of neural machine architecture, a big-data is always required to get a good performance.
To achieve that, a dialogue set was crawled as the training set.
And to be compared with,
a basic kind of Seq2Seq architecture for response generation called hybrid model
proposed by \newcite{Shang2015Neural} was implemented.
\subsection{Data}
For the training process, Some one-round dialogue pairs was crawled from the
Internet. For convenience, first sentence and the second
sentence of one dialogue pair are denoted as post and response\cite{Shang2015Neural}respectively.
The data set contains one million
pairs, and about 35 thousands words.
It should be noted
that compared with the data used by \newcite{Shang2015Neural},
this crawled data is a one-to-one data set, one post is corresponding
to exact one response. While in the \newcite{Shang2015Neural} paper, they crawled some
one-to-many data from microblog, then distributed all the responses
to the its post. This is a creative way to build a big data set, while
during our experiments, we found that the one-to-one data has a
more rapid rate of convergence, so we created our own data set and
trained models on it.Table \ref{trainingdata} is an example of our data.
\begin{table}[h]\small
\begin{center}
\begin{tabular}{|l|l|}
\hline post & response \\ \hline
\tabincell{l}{今天天气好差呀\\The weather is so bad today} & \tabincell{l}{雨太大了\\The rain is too heavy} \\ \hline
\tabincell{l}{每天六点多出去打篮球 锻炼身体\\ Go out for exercise playing basketball at six every day}& \tabincell{l}{我在打网球 \\I am playing tennis}\\ \hline
\tabincell{l}{白色搭配什么颜色好\\What color matches white well} & \tabincell{l}{白色百搭呀 \\White all-match} \\\hline
\tabincell{l}{明晚我又要通宵\\I will stay night again tomorrow night} & \tabincell{l}{我陪你啦 \\I will be with you} \\ \hline
\tabincell{l}{等我有空了去超市买\\I'll go to the supermarket to buy when I was free} & \tabincell{l}{超市太远了\\The supermarket is too far away}\\
\hline
\end{tabular}
\end{center}
\caption{\label{trainingdata} Training data examples. }
\end{table}

As for the test set, considering that one of our evaluation method--Bleu,
which will be introduced in detail in next section, should has more than one
reference for every candidate, our one-to-one data is not very suitable,
so we select 100 posts and their corresponding responses in \newcite{Shang2015Neural}'s data-set
to build our test set.  Table \ref{data set information} shows some statics
of our the whole data set.
\begin{table}[htbp]\small
\begin{center}
\begin{tabular}{|c|c|c|c|}
\hline Data & Data type & posts & responses \\ \hline
Training Data & one-to-one &  1000,000 & 1000,000 \\ \hline
Test Data & one-to-many & 1000 & 42422 \\ \hline
\end{tabular}
\end{center}
\caption{\label{data set information}Data statistics}
\end{table}
\vspace{-1em}
\subsection{Models}
We trained two models. The first one is
a basic Seq2Seq model for dialogue called Hybird Model(denoted as HYP)\cite{Shang2015Neural} ,
the other is what we have proposed(denoted as LTS).
The encoder and decoder were both implemented using the GRU \cite{Cho2014On,Chung2014Empirical}.
and we set our model's parameters reference to \newcite{Shang2015Neural}.
The hidden size in the encoder was set to 1024. And the embedding size was set to 500,
all the embedding vectors were pre-trained using the training data \cite{mikolov2013efficient,mikolov2013distributed}.

Besides, during the processing of training, we sent the data to the model using the mini-batch
with a batch size of 100, and the RMSprop algorithm was used to update model's parameters.
And we trained both models about 5 days. After that, we used the beam search algorithm to search for the N-best
result of response for one paritular sentence \cite{graves2012sequence,boulanger2013audio}. 

\section{Result}
Until now, there is still not a uniform evaluation for response generation \cite{galley2015deltableu,pietquin2013survey,schatzmann2005quantitative}.First, we tested our model's performance using the some statistics of the the first predicted word.
Second, we evaluate the complete response to see if our model can bring the Seq2Seq architecture improvement.
to achieve that goal, we use two metrics: for one hand, we employed the wildly
used automatic evaluation method--blue \cite{papineni2002bleu} in the area of machine translation,
for the other hand, we employed the human annotation method.

\subsection{First Generated Word Evaluation}
To evaluate the generation of the first word, two aspects are taken into consideration: the accuracy rate and the diversity.
The dialogue pairs in the test set are denoted as test sample and reference respectively, and
the sequence generated by the model is still denoted as response.
As mentioned before, the test set is a one-to-many data set so each test sample
corresponding to serval references. We define a set called R-set for every sample, each sample's R-set
is composed by all the first words of that sample's references, during the test process,
if the first word of sample's response fall into its R-set, then it will be marked as hit. And the
accuracy is the ratio of hit samples over the whole test set.
Furthermore, we considered
such situation: some high-frequency words(derived from the training data) are so common that nearly all the R-sets consists at least one,
so a sample will easily hit its R-set as well as its first generated word is such words, for example:'I'.
So we further defined the accuracy without high-frequency words, denoted by $accw$-$i$,which takes such
situation into consideration:
if a hit word is one of top $i$ high-frequency words, then this hit will be ignored.
Particularly, $accw$-$0$ equals the basic accuracy that do not filter any high-frequency words.

\begin{figure}[!htb]	
	\centering
    \includegraphics[width=0.5\textwidth,natwidth=610,natheight=642]{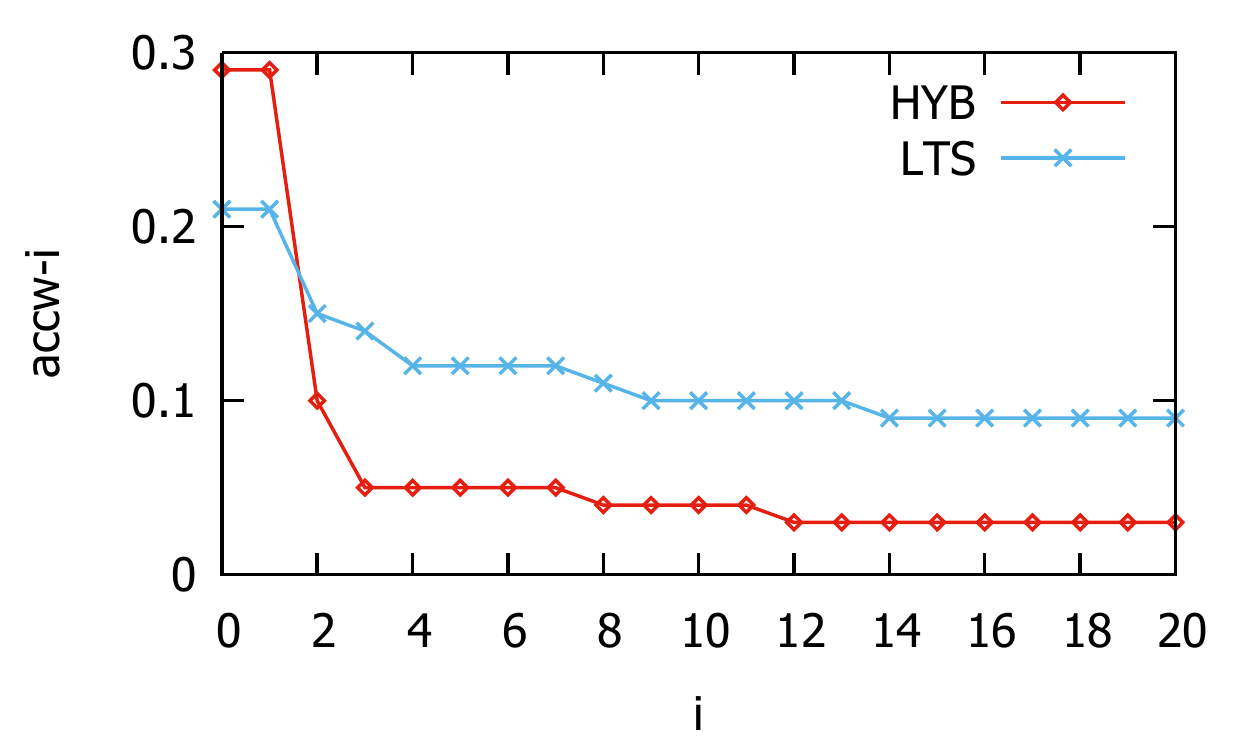}\\
	\caption{The accw-i metric}
	\label{acc}
\end{figure}
\vspace{-1cm}

From the Figure \ref{acc}, we can see that the LTS outperformed the HYP from the $accw$-$2$.We analyzed
the results and find the most frequent word is a auxiliary word: "了", which seldom appear in the beginning
of a sentence, so there is no change from the $accw$-$0$ to $accw$-$1$.When we ignored the second frequent
word "我(I)", the performance of the HYP descends rapidly,which can be observed from the $accw$-$1$ to $accw$-$2$.
while the LTS has a more stable accuracy that do not depend the easily hit high-frequency words.

Also, we evaluate the initial prediction from the perspective of diversity. In fact the Table \ref{acc}
has already reflected the diversity to some degree, which our model's stable $accw$-$i$ shows that the generation of the first word is distributed fairly balance. While we still give another metric to evaluate it, we define
the div-i which means the ratio of test samples whose response's first word fall into the top $i$
frequent words.According to the definition we describe, we can see that the diversity declines with increasing div-i score.

\begin{figure}[!htb]	
	\centering
    \includegraphics[width=0.5\textwidth,natwidth=610,natheight=642]{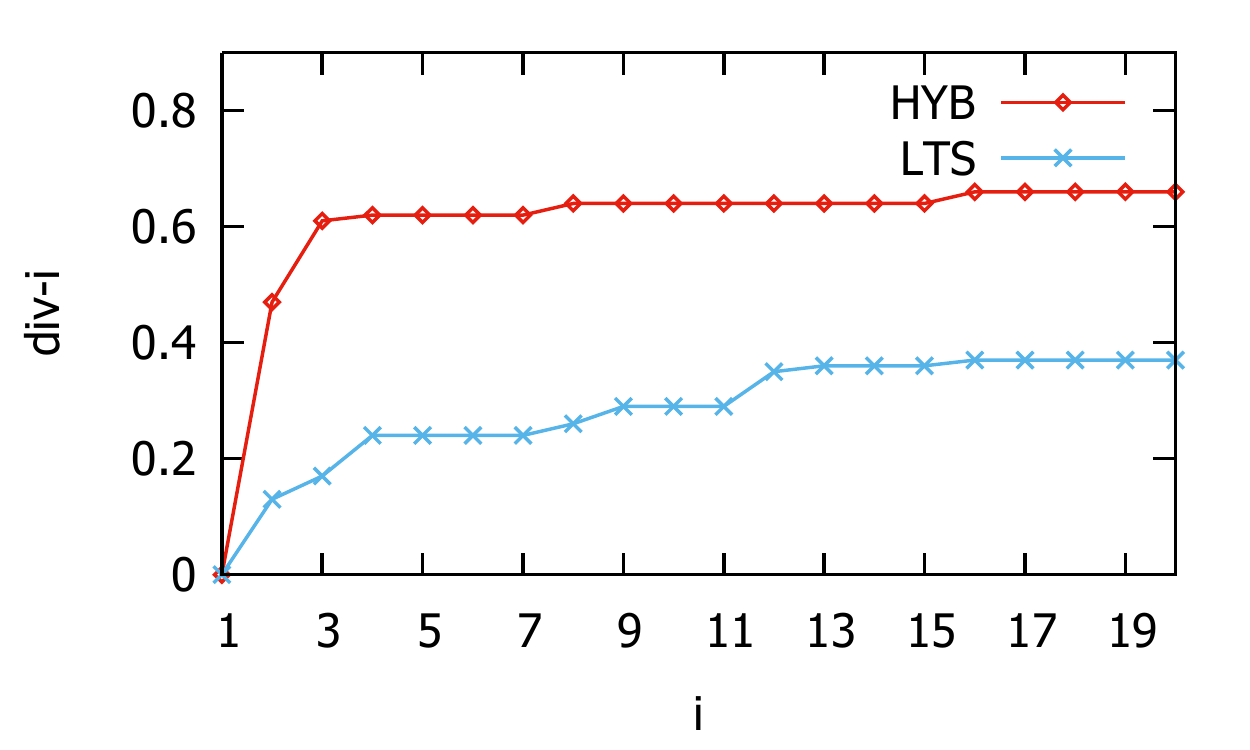}\\
	\caption{The div-i metric.}
	\label{div}
\end{figure}

The Figure \ref{div} show us that rather than concentrate on some high-frequency words, our model prefer
to predict more diversity ones. Noted that in HYP there is a sharp increasement from
the $div$-$1$ to $div$-$2$, which indicates that lots of the samples's first generated word is the second frequent
word, which also agrees the results of $acc$-$i$.

\subsection{Bleu Metric}
We use this metric to evaluate the a model's complete response rather than the first word.
which is proved to agree well with human judgement on
response generation task \cite{sordoni2015neural,li2015diversity}.
And the result is given in the Table \ref{bleu}.

\begin{table}[!htb]\small
\begin{center}
\begin{tabular}{|c|c|c|c|c|}
\hline & BLEU-1 & BLEU-2 & BLEU-3 \\ \hline
HYP & 0.5283 & 0.0553 & 0.0013 \\ \hline
LTS & \textbf{0.5303} & \textbf{0.0816}& \textbf{0.0063} \\
\hline
\end{tabular}
\end{center}
\caption{\label{bleu} Bleu score. }
\end{table}
From the Table \ref{bleu}, we can see that the LTS performs well than the HYP
in Bleu-1 to Bleu-3. Through that table, we can also see that improvement on the Bleu-1 is
not as significant as other two. We analyzed this situation and got an opinion, it may because
the Bleu metric calculate overlap of n-grams between response and references, compared with
other n-grams, the unigram is more easily to be matched making the Bleu-1 not distinguished
enough.Table \ref{bleu example} shows some results of the bleu evaluation,
two models got similar Bleu-1 scores, while the Bleu-2 and Bleu-3 are much more strict metrics
that can reflect the improvement more significantly, which also proves our analyzed mentioned before.

\begin{table}[!htb]\small
\begin{center}
\renewcommand{\multirowsetup}{\centering}
\begin{tabular}{|c|l|l|l|c|c|c|}
\hline No. &model& post& responds & Bleu-1 & Bleu-2 &Bleu-3 \\ \hline
\multirow{2}{*}{1} & HYP & \multirow{2}{*}{\tabincell{l}{拍照中，这个造型像个娃娃一样！\\ Taking pictures, this pose looks\\ like a little doll!}} & \tabincell{l}{不像男的\\ Doesn't like a man} & 0.750 & 0.000 & 0.000 \\ \cline{2-2} \cline{4-7}
&LTS & & \tabincell{l}{都是小娃娃\\ All are little dolls}  & 1.000 & 0.500 & 0.167 \\
\hline
\multirow{2}{*}{2} & HYP & \multirow{2}{*}{\tabincell{l}{低调的境界\\ The state of low-key}} & \tabincell{l}{我也很低呀\\ I'm very low,too} & 0.250 & 0.000 & 0.000 \\ \cline{2-2} \cline{4-7}
&LTS & & \tabincell{l}{我很低调的\\ I am very low-key}  & 0.800 & 0.400 & 0.133 \\
\hline
\end{tabular}
\caption{\label{bleu example} Bleu evaluation examples. }
\end{center}
\end{table}

\subsection{Manual Evaluation}
At the same time, we also tested our model adopting human annotation method. The evaluation
metric is made reference to Shang~\shortcite{Shang2015Neural}. We generated responses with HYP
model and LTS respectively, then these responses together with their original
questions are mixed up into a new file to make sure that labers can judge the result fairly.
Three labers were
involved to assigned a score to these responses in range of 0 to 2, and the score metric
is as follows:

0:This indicates a bad response.
 if a response has grammar,fluency mistakes, not logic consist or relevant with the original post in semantic, it should be assigned a 0 score.

1:This means the response may not be a perfect one, but may be treated as a
suitable response in some particular scenario, or it is a too general response, like "I don't know".

2:This indicates a quite appropriate response, a response can be sorted to
this category only when it is free of grammar and fluency errors and is independent of
scenario.

\begin{table}[!htb]\small
\begin{center}
\begin{tabular}{|c|l|l|c|}
\hline No.& post & response & score\\ \hline
1 & \tabincell{l}{咳嗽不止，有啥良策吗\\ Couldn't stop cough, any good idea?} & \tabincell{l}{吃点退烧药 \\eat some antipyretics} &0 \\ \hline
2 & \tabincell{l}{永远不要对一个外行聊你的专业 \\ Never talk about your major with strangers} & \tabincell{l}{我很专业的 \\ I'm very professional} &0\\ \hline
3 & \tabincell{l}{真实的团队精神是这样子的~ \\This is what real team spirit likes ~}& \tabincell{l}{是挺厉害的 \\
It's pretty powerful.} &1\\ \hline
4 & \tabincell{l}{哪本新华字典是你用过的 \\ which XinHua dictionary is the one you used} & \tabincell{l}{两本都是\\Both of them are} &1\\ \hline
5 & \tabincell{l}{大熊加油，我们的家) \\Come on for our home, Bill}& \tabincell{l}{我会努力的\\I'll try my best} &2\\
\hline
\end{tabular}
\end{center}
\caption{\label{metric} Annotation metric examples. }
\end{table}

\begin{table}[h]\small
\begin{center}
\begin{tabular}{|c|c|c|c|c|c|}
\hline Models & Mean Score & 0 & 1 & 2 & Agreement \\ \hline
HYB & 0.510 & 66.0\% & 17.0\% & 17.0\% & 0.230 \\\hline
LTS & \textbf{0.590} & \textbf{59.7\%} & \textbf{21.7\%} & \textbf{18.6\%} & 0.206\\ \hline
\end{tabular}
\end{center}
\caption{\label{font-table} Annotation result. }
\end{table}
\vspace{-1em}
Table \ref{metric} show the annotation metric in more detail. The example one conflicts
with the logic consistency principle, the post said he got a cough, while the response
advised the antipyretics, which is not logical relevant. The example two's response is not
semantic relevant to its post, so they got a score
of zero. In the example three, the response can be seen as a suitable one but it is
too general, so it only got a score of one. While the example four got the same score in a
different way that its response strongly depend on a particular scenario that the author
of the post must has exactly two dictionaries. The last example show a suitable response that
free of the questions that mentioned before.

The human annotation result was used to
compute these metrics for the two model respectively: mean score, ratios
of different categories.
From the table 4, we can see that the LTS outperforms the HYP in all
metrics. Besides, we evaluated the consistent of different labelers using
the Fleiss' kappa, which is listed the Agreement column. We can see that
both of the two models' agreement all fall into the range of 0.2-0.4,
which indicates that the result is a fair agreement one.

\begin{table}[!htb]\small
\begin{center}
\begin{tabular}{|c|l|l|l|}
\hline No. & post & HYP & LTS \\ \hline
1 & \tabincell{l}{你们听说过搜狗输入法石化版吗？\\Have you heard of of sogou pinyin's petrochemical\\ version} & \tabincell{l}{这是国服\\This is National Server} & \tabincell{l}{听说挺好的\\I heard it's nice}\\ \hline
2 & \tabincell{l}{别管我为什么叫卢沟桥\\Don't worry about me why called Marco Polo bridge} & \tabincell{l}{那叫什么啊\\What's it called?} & \tabincell{l}{谁管你了\\Who cares for} \\ \hline
3 & \tabincell{l}{这才是大学最大的遗憾\\This is the greatest regret in university} & \tabincell{l}{我也怕大学\\I also afraid of university} & \tabincell{l}{有啥遗憾的\\There is nothing to regret} \\ \hline
4 & \tabincell{l}{“师兄帮帮忙网站”城市开放计划。\\“Ask Brother for Help websitde” Urban open plan}& \tabincell{l}{我很开放的\\I am very open} & \tabincell{l}{很好的项目\\It's a good prodject} \\ \hline
5 & \tabincell{l}{现在开会需要用上喇叭扩音了……\\Now the meeting need to use the\\ horn to be heard……} & \tabincell{l}{不需要补的\\Don't need to fill} & \tabincell{l}{没钱买喇叭\\Don't have the money\\ to buy a horn} \\ \hline
\end{tabular}
\end{center}
\caption{\label{results} Some results of two models. }
\end{table}
\vspace{-1em} 

\section{Related Work}
\subsection{Sequence to sequence for Machine Translation}
Using the sequence to sequence model, neural machine translation has already
got a comparable performance to the traditional methods \cite{bahdanau2014neural}.
As far as we know, it was first introduced into this area by \newcite{kalchbrenner2013recurrent},\newcite{Sutskever2014Sequence},\newcite{Cho2014Learning},\newcite{gao2014learning}.
Besides, \newcite{Cho2014Learning} added the vector c
as an extra input to every time step of the decoder, by doing like this, all
the steps not only the first one, can get full information of the context vector.
Furthermore, \newcite{bahdanau2014neural} proposed a novel method to calculated a weighted sum of all the annotations
of the encoder. This mechanism can be
regarded as a kind of attention, which means when we decode a word we chose
which part of the annotations should be paid more attention to.

\subsection{Sequence to sequence in Response Generation}
General speaking, dialogue systems can be sorted into two classes \cite{serban2016building}:
goal-driven represented by systems \newcite{gavsic2013line} and non-goal-driven systems.
The neural networks methods are mainly used in the later, because a large
scale of data is more easily to get in that area.
\newcite{ritter2011data} first combine micro-blogging data
with the generative probabilistic models, then
\newcite{Shang2015Neural} used this type of data on the Seq2Seq to build
a short conservation machine. followed by \newcite{serban2016building}, who came up with the Hierarchical Nerual
Network model, aiming to model the utterances and interactive structure to build
a multi-round dialogue system.

At the same time, \newcite{banchs2012movie} proposed methods using a different type of data,the
movie dialogue. Based on that, \newcite{ameixa2014luke} find using the retrieal system and movie
subtitles can also improvement the performance.

\section{Conclusions}
In this paper, we proposed a new approach for the sequence to sequence model to generate the first word.
Proved our proposed model can bring a promotion in both the accuracy and the diversity for the first word's generation,
thus improving the whole performance of the generation.
Experiments in the response generation tasked verified our model's effectiveness,
while rather than a method for a specific task, our proposed
method is a general framework, which can also used for other tasks. 

\bibliographystyle{acl}
\bibliography{coling2016}

\end{CJK*}
\end{document}